\documentclass[10pt,twocolumn,letterpaper]{article}

\usepackage{iccv}
\usepackage{times}
\usepackage{epsfig}
\usepackage{graphicx}
\usepackage{amsmath}

\usepackage{amssymb}

\usepackage{graphics} 
\usepackage{epsfig} 
\usepackage{mathptmx} 
\usepackage{times} 
\usepackage{amsmath} 
\usepackage{amssymb}  
\usepackage{multirow}

\usepackage{enumerate}
\usepackage{float}
\usepackage{siunitx} 
\usepackage{algpseudocode}
\usepackage{algorithm}
\usepackage {xcolor}

\usepackage{times}
\usepackage{xspace}
\usepackage{xcolor}
\usepackage{booktabs }
\usepackage{enumitem}
\usepackage{makecell}
\usepackage{comment}

\newcommand{\B}[1]{\textcolor{blue}{#1}}

\newcommand{\figref}[1]{Fig\onedot~\ref{#1}}
\newcommand{\equref}[1]{Eq\onedot~\eqref{#1}}

\newcommand{\tabref}[1]{Tab\onedot~\ref{#1}}

\newcommand{\ALGref}[1]{Alg\onedot~\ref{#1}}

\newcommand{\thickhline}{%
	\noalign {\ifnum 0=`}\fi \hrule height 1pt
	\futurelet \reserved@a \@xhline
}

\newcommand{\PreserveBackslash}[1]{\let\temp=\\#1\let\\=\temp}
\newcolumntype{C}[1]{>{\PreserveBackslash\centering}p{#1}}
\newcolumntype{R}[1]{>{\PreserveBackslash\raggedleft}p{#1}}
\newcolumntype{L}[1]{>{\PreserveBackslash\raggedright}p{#1}}

\usepackage[pagebackref=true,breaklinks=true,letterpaper=true,colorlinks,bookmarks=false]{hyperref}

 \iccvfinalcopy 

\def\iccvPaperID{2261} 

\ificcvfinal\fi
\begin{document}

\title{Augmented LiDAR Simulator for Autonomous Driving}

\author{Jin Fang, Dingfu Zhou,Feilong Yan, Tongtong Zhao, Feihu Zhang, Yu Ma and Liang Wang, Ruigang Yang
\\ \{fangjin, zhoudingfu,yanfeilong, zhaotongtong01, zhangfeihu, mayu01, wangliang18, yangruigang\}@baidu.com
}
\maketitle
\thispagestyle{plain}
\pagestyle{plain}
\begin{abstract} 
    
In Autonomous Driving (AD), detection and tracking of obstacles on the roads is a critical task. Deep-learning based methods using annotated LiDAR data have been the most widely adopted approach for this. 
Unfortunately, annotating 3D point cloud is a very challenging, time- and money-consuming task. In this paper, we propose a novel LiDAR simulator that augments real point cloud with synthetic obstacles (e.g., cars, pedestrians, and other movable objects). Unlike previous simulators that entirely rely on CG models and game engines, our augmented simulator bypasses the requirement to create high-fidelity background CAD models. Instead, we can simply deploy a vehicle with a LiDAR scanner to sweep the street of interests to obtain the background point cloud, based on which annotated point cloud can be automatically generated. This unique ''scan-and-simulate" capability makes our approach scalable and practical, ready for large-scale industrial applications. In this paper, we describe our simulator in detail, in particular the placement of obstacles that is critical for performance enhancement. We show that detectors with our simulated LiDAR point cloud \textbf{alone} can perform comparably (within two percentage points) with these trained with real data. Mixing real and simulated data can achieve over 95\% accuracy. 

\end{abstract}

\section{Introduction}


\begin{figure}[ht!]
	\centering
	\includegraphics[width=0.50\textwidth]{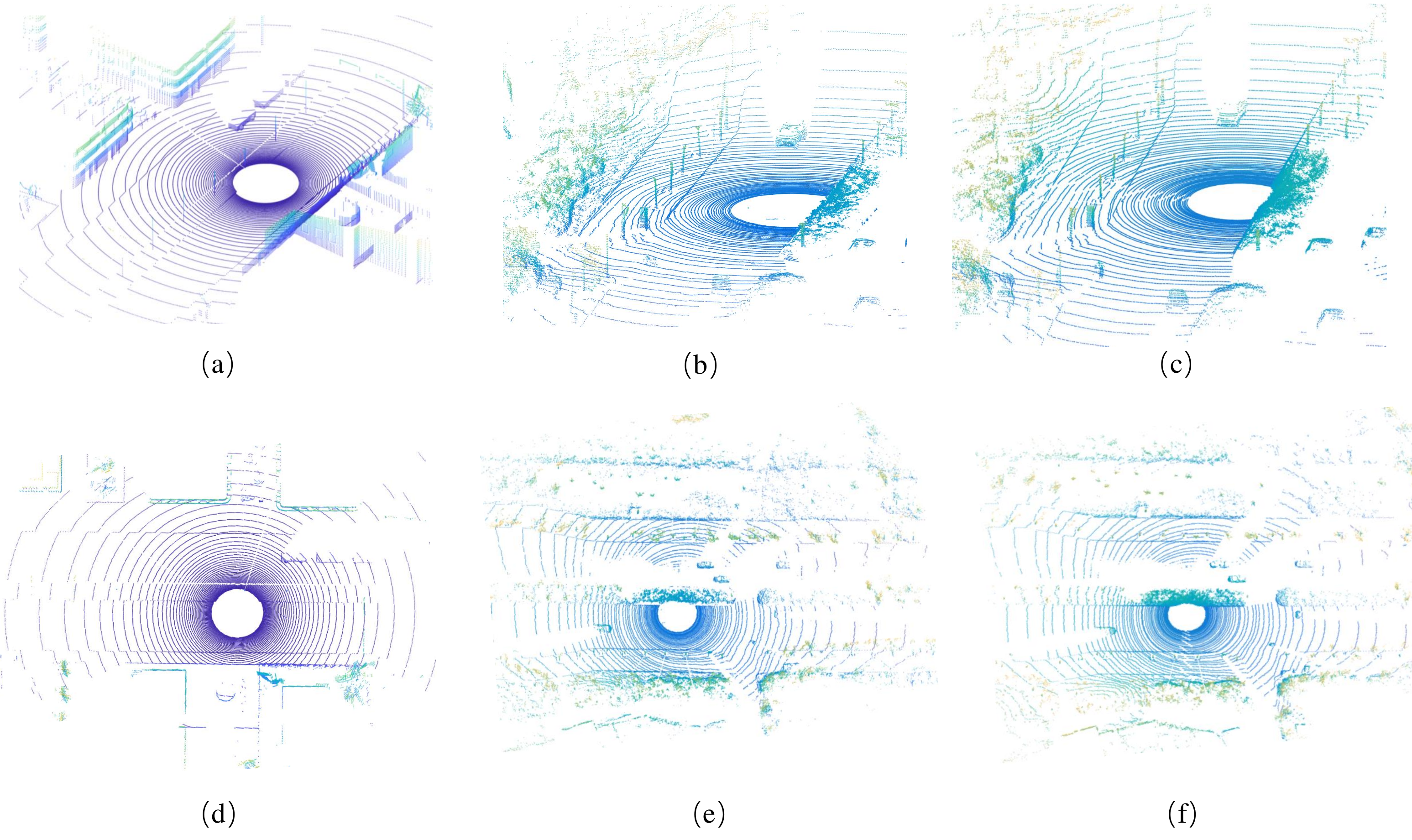} 
	\centering
	\caption{Simulation point cloud with different methods. (a) is generated from CARLA, (b) comes from our proposed method and (c) is the real point cloud collected by Velodyne HDL-64E. The second row shows the point cloud from the bird-eye-view. Notice the inherently rich background in our approach.}
	\label{Fig:pipe_framework}
\end{figure}

LiDAR devices have been widely used in robotics and in particular autonomous driving. They provide robust and precise depth measurements of their surroundings, making them usually the first choice for environmental sensing. Typically the raw point cloud from LiDAR is sent to a computer vision system to find obstacles and other relevant driving information. Currently high-performance vision systems usually are based on deep learning techniques. Deep neural networks (DNN) have proven to be a powerful tool for many vision tasks \cite{chen2017multi} \cite{zhou2018voxelnet} \cite{xu2018pointfusion}.

The success of DNN most relies on the quality and quantity of labeled training data. Compared to image data, 3D point cloud from LiDAR is much more difficult to label manually\cite{kopper2011rapid} \cite{welinder2010online}. This is particularly true from real-time LiDAR scanners, such as these from Velodyne. These devices, predominantly used in autonomous vehicles (AV),  generate sparse point cloud that is difficult to interpret. Labeling the huge amount of point cloud data needed for the safety of AV quickly becomes prohibitively expensive. 

There have been work to leverage computer graphics techniques to generate synthetic labeled data (e.g.,~\cite{su2015render}, \cite{richter2017playing}, \cite{mueller2017ue4sim}, \cite{wu2018squeezeseg}, \cite{wu2018squeezesegv2}).  
While these simulated data are shown to be useful to improve DNN's performance, there remain a few unsolved problems. First the CG environment is mostly manually crafted with limited scale and complexity. As we will show in our experiments, the fidelity of background plays a significant role in perception accuracy. Creating photo-realistic scenes, with a price tag of over 10K USD per kilometer, simply does not scale. Secondly the obstacle placement and movement are mostly based on heuristics, which often do not reflect the diversity of our real world. Thirdly in the scope of LiDAR simulation~\cite{yue2018lidar} \cite{wu2018squeezesegv2}, existing methods simply render the scene depth, without considering the physical characteristics of LiDAR, leading to obvious artifacts. As such detector trained with only synthetic data performed poorly (e.g., around $30\%$ accuracy) on real data~\cite{yue2018lidar}. 

In this paper we present a novel hybrid point cloud generation framework for automatically producing high-fidelity annotated 3D point data, aimed to be immediately used for training DNN models. To both enhance the realism of our simulation and reduce the cost, we have made the following design choices. First, we take advantage of mobile LiDAR scanners, which are usually used in land surveys, to directly take 3D scan of road scenes as our virtual environment, which naturally retains the complexity and diversity of real world geometry, therefore bypassing completely the need for environmental model creation. Secondly we develop novel data-driven approaches to determine obstacles' poses (position and orientation) and shapes (CAD model). More specifically, we extract from real traffic scenes the distributions of obstacle models and their poses. The learned obstacle distribution is used to synthesize the placement and type of obstacles to be placed in the synthetic background. Note that the learning distribution does not have to be aligned with the captured background. Different combinations of obstacle distribution and background provide a much richer set of data without any additional data acquisition or labeling cost.  Thirdly we develop a novel LiDAR renderer that takes into considerations both the physical model and real statistics from the corresponding hardware. Combing all these together, we have developed a simulation system that is realistic, efficient, and scalable.


The main contributions of this paper include the following. 
 \begin{itemize}
	\item  We present a LiDAR point cloud simulation framework that can generate the annotated data for autonomous driving perception, the resultant data have achieved comparable performance with real point cloud.  Our simulator provides the realism from real data, with the same amount of flexibility that was previously available only in VR-based simulation, such as regeneration of traffic patterns and change of sensor parameters.
	
	\item We combine real-world background models, which is acquired by LiDAR scanners, with realistic obstacle placement that is learned from real traffic scenes. Our approach does not require the costly background modeling process, or heuristics based rules. It is both efficient and scalable.  
	\item We demonstrate that the model trained with synthetic point data alone can achieve competitive performance in terms of 3D obstacle detection and semantic segmentation.  Mixing real-data and simulated data can easily outperform the model trained with the real data alone.  
\end{itemize}

 \section{Related Work} \label{sec:related_work}
As deep learning becomes prevalent, increasing effort has been invested to alleviate the lack of annotated data for training DNN. In this section, we mainly review the recent work on data simulation and synthesis for autonomous driving.

To liberate the power of DNN from limited training data, \cite{ros2016synthia} introduces SYNTHIA, a dataset of a big volume of synthetic images and associated annotation of urban scenes. \cite{gaidon2016virtual} imitates the tracking video data of KITTI in virtual scenes to create a synthetic copy, and then augments it by simulating different lighting and weather conditions. \cite{gaidon2016virtual} and \cite{richter2017playing} generate a comprehensive dataset with pixel level labels and instance level annotation, aiming to provide a benchmark supporting both low-level and high-level vision tasks. \cite{cheung2018mixedpeds} added CG characters to existing street view images for the purposed of pedestrians detection. However the problem of existing pedestrians in the image is not discussed. \cite{johnson2017driving} synthesizes annotated images for vehicle detection, and shows an encouraging result that it is possible for a state-of-the-art DNN model trained with purely synthetic data to beat the one trained on real data, when the amount of synthetic data is sufficiently large. Similarly, our target is to obtain the comparable perception capability from purely simulated point data by virtue of their diversity. To address the scarcity of realism of virtual scenes,\cite{alhaija2018augmented} proposes to augment the dataset of real-world images by inserting rendered vehicles into those images,  thus inheriting the realism from the background and taking advantage of the variation of the foreground. \cite{tsirikoglolu2017procedural} presents a method for data synthesis based on procedural world modeling and more complicated image rendering techniques. Our work also benefits from the real background data and physically based sensor simulation.

While most of previous work are devoted to image synthesis, only few focus on the generation and usage of synthetic LiDAR point cloud, albeit they play an even more important role for autonomous driving. Recently, \cite{yue2018lidar} collects the calibrated images and point cloud using the APIs provided by the video game engine, and apply these data for vehicle detection in their later work \cite{wu2018squeezeseg}. Carla \cite{dosovitskiy2017carla} and AutonoVi-Sim \cite{best2017autonovi} also furnish the function to simulate LiDAR point data from the virtual world. However their primary target is to provide platform for testing algorithms of learning and control for autonomous vehicles. 

In summary, in the domain of augmenting data,  our method is the first to focus on LiDAR point. In addition, we can change the parameters of LiDAR, in terms of placement and the number of lines, arbitrarily. The change of camera parameters has not been reported in image-augmentation methods.

\begin{figure*}[ht!]
	\centering
	\includegraphics[width=1.0\textwidth]{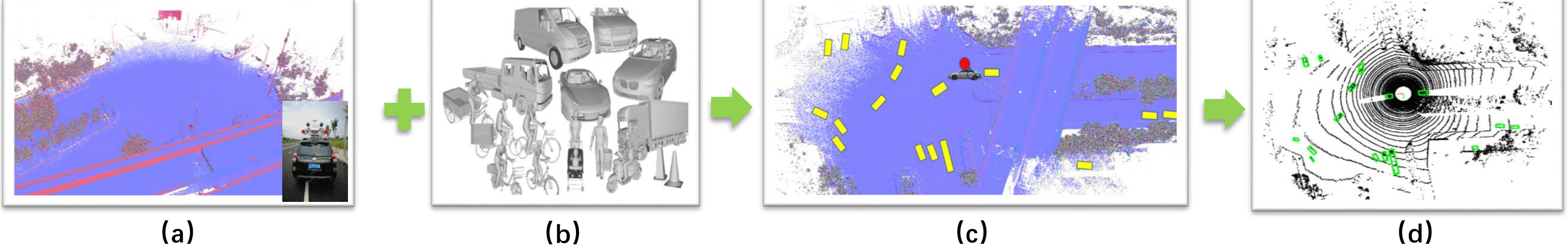} 
	\centering
	\caption{The proposed LiDAR point cloud simulation framework. (a) describes accurate, dense background with semantic information obtained by a professional 3D scanner. (b) shows the synthetic movable obstacles e.g., vehicle, cyclist and other objects. (c) illustrates an example of placing foreground obstacles (yellow boxes) in the static background based on a Probability Map. (d) is an example of the simulated LiDAR point cloud with ground truth 3D bounding boxes (green boxes) by using our carefully designed simulation strategy.  }
	\label{Fig:pipe_framework}
\end{figure*}

\section{Methodology}
In general, we simulate the data acquisition process of the LiDAR sensor mounted on the autonomous driving vehicle in the real traffic environment. The whole process is composed of several modules: static background construction, movable foreground obstacles generation and placement, LiDAR point cloud simulation and the final verification stage. A general overview of our proposed framework is described in \figref{Fig:pipe_framework} and more details of each module will be described in the following context.

\subsection{Static Background Generation} \label{subsec:background}

 Different with other simulation frameworks \cite{dosovitskiy2017carla} \cite{yue2018lidar}, which generate both the foreground and background point cloud from an artificial virtual world, we generate the static background with the help of a professional 3D scanner Riegl VMX-1HA \footnote{Riegl VMX-1HA \url{http://www.riegl.com/nc/products/mobile-scanning/produktdetail/product/scanner/52/}}.  
 
\begin{figure}[ht!]
	\centering
	\includegraphics[width=0.5\textwidth]{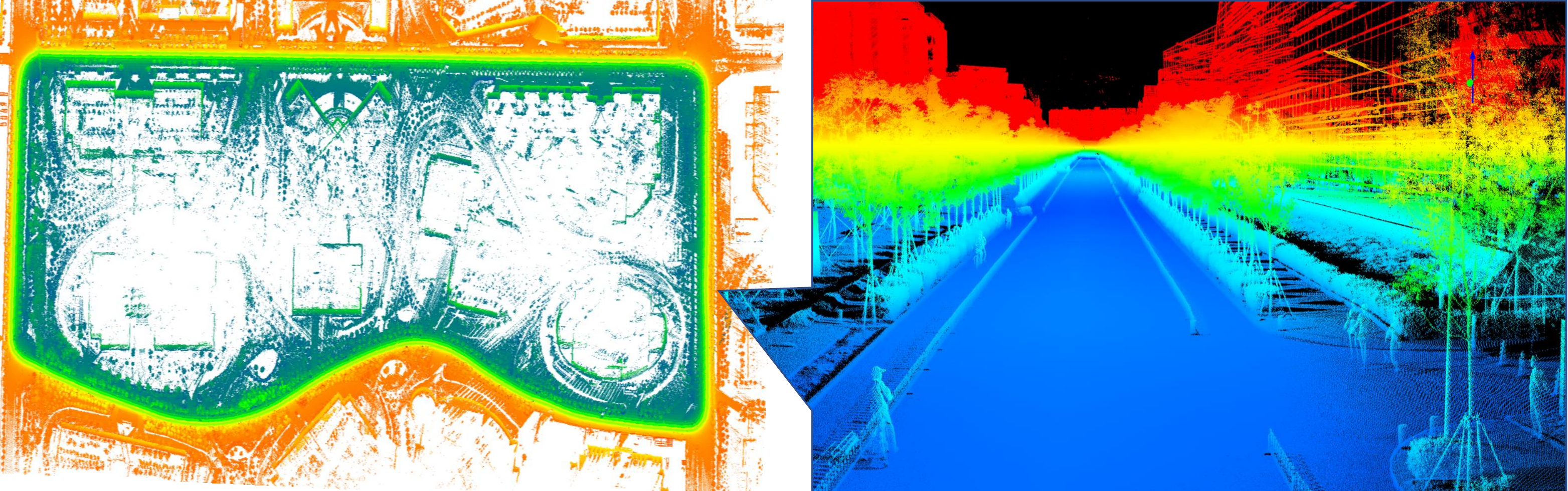} 
	\centering
	\caption{Left sub-image illustrates an example of the point cloud obtained by RIEGL scanner with more than 200 million 3D points. The actual size of the place is about \SI{600}{\m} $\times$ \SI{270}{\m}. Right sub-image displays the detail structure of the point cloud.}
	\label{Fig:RIGEL_pointcloud}
\end{figure}

The RIEGL is a high speed, high performance dual scanner mobile mapping system which provides dense, accurate, and feature-rich data at highway speeds. The resolution of the point cloud from the Riegl scanner is about \SI{3}{\cm} within a range of 100 meters. In the real application, a certain traffic scene will be repeatedly scanned several rounds (e.g., $5$ rounds) in our experiments. After multiple scanning, the point cloud resolution can be increased to about \SI{1}{\cm}. An example of the scanned point cloud is displayed in Fig \ref{Fig:RIGEL_pointcloud}. By using this scanner, the structure details can be well obtained. Theoretically, we can simulate any other type of LiDAR point cloud whose point distance is larger than \SI{1}{\cm}. For example, the resolution of common used Velodyne HDL-64E S3 \footnote{Velodyne HDL-64E S3 \url{https://velodynelidar.com/hdl-64e.html}} is about \SI{1.5}{\cm} within a range of 10 meters. 
 
To obtain clean background, both the dynamic and static movable obstacles should be removed away. To improve the efficiency, state-of-the-art semantic segmentation approaches are employed to obtain an initial labeling results roughly and then annotators correct these wrong parts manually. Here, PointNet++ \cite{qi2017pointnet++} is used for semantic segmentation and the average precision can reach about $94.0\%$. Based on the semantic information, holes are filled with the surrounding points.

 \subsection {Movable Obstacle Generation}
\label{subsec:obs-dis}
After obtaining the static background, we need to consider how to add movable obstacles in the environment. Particularly, we find that the position of obstacles has a great influence on the final detection and segmentation results. However, this has been rarely mentioned by other simulation approaches. Instead of placing obstacles randomly, we propose a data-driven-based method to generalize the obstacle's pose based on their distribution in the real dataset. 
\subsubsection{Probability Map for Obstacle Placement} First of all, a Probability Map \cite{cheung2018mixedpeds} will be constructed based on the obstacles distribution in the labeled dataset from different scenarios. In the probability map, the position with a higher value means it will be selected to place an obstacle with a higher chance. This map is built based on some labeled dataset. Instead of merely increasing the probability value at the position where an obstacle appeared in the labeled dataset, we also increase the neighboring positions based on a Gaussian kernel. Similarly, the direction of the obstacles can also be generated from this map. Details of building probability map can be found in \ \ALGref{alg:Probability_map}. Particularly, we have built different probability maps for different classes. Given the semantics of background, we could easily generalize the Probability Map to other areas in a texture-synthesis fashion.
\begin{algorithm}[ht!] 
\caption{Probability Map based Obstacle Pose Generation}
\begin{algorithmic}[1]
\Require{\begin{minipage}[t]{1\textwidth}
- Annotated point cloud $\mathbf{S}$ in a local area\\
- A scanner pose $\mathbf{p}$;
\end{minipage}}
\Ensure{
~- A set of obstacles' pose; 
}
\noindent \begin{raggedright}
\rule[0.15dd]{0.999\linewidth}{1pt}
\par\end{raggedright}
\State $\blacktriangleright\;$Divide the local area $\mathbf{G}$ into $M \times N$ grids with the weight matrix $\mathbf{W}$ and direction matrix $\mathbf{\theta}$;
\State $\blacktriangleright\;$Initialize the Gaussian weight template $\mathbf{T}$ with the size of $(2k+1) \times (2k+1)$;
 \For{$i,j \gets 1$ to $M,N$}     
         \If{$\mathbf{G}_{i,j}$ contain obstacles}
          \For{$m,n \gets -k$ to $k$}     
                \State  $\mathbf{W}_{i+m,j+n} \:\:\: +=
                \mathbf{T}_{m,n}$
                 \State  $\mathbf{\theta}_{i+m,j+n} \quad +=  
                \mathbf{W}_{i+m,j+n}*\mathbf{\theta}_{i,j}$
           \EndFor
         \EndIf
 \EndFor

\State $\blacktriangleright\;$ Given a scanner pose $\mathbf{p}$, obstacle positions and directions can be sampled with $\mathbf{W}$ and $\mathbf{\theta}$ by weighted random sampling; 
\end{algorithmic}\label{alg:Probability_map}
\end{algorithm}
\normalsize



\begin{figure}[ht!]
	\centering
	\includegraphics[width=0.5\textwidth]{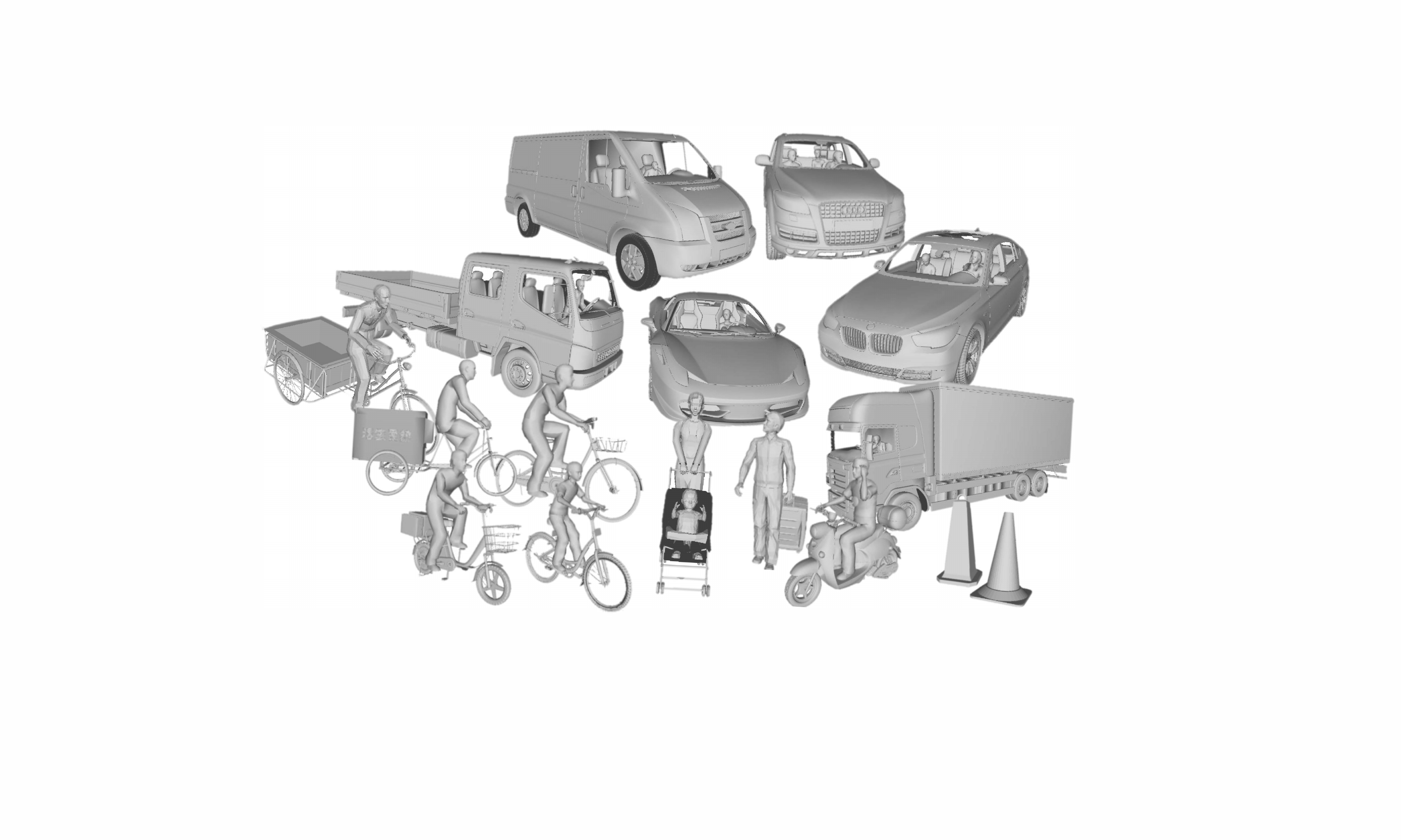} 
	\centering
	\caption{Man-made CAD models. For real AD application, some uncommon categories have been also considered in our simulation, e.g., traffic cone, baby carriage and tricycle, etc.}
	\label{Fig:CAD_Modes}
\end{figure}

\subsubsection {Model Selection} 
Similar to the obstacle's position, a data-driven-based strategy has been employed to determine the occurrence frequency of different obstacle categories. Based on the labeled dataset, prior occurrence frequency information of different types can be easily obtained. During the simulation process, each 3D model will be selected with this prior knowledge. In addition, the model for each category has been divided into two groups: one is the high-frequency model used to cover the most common cases and another is the low-frequency model used to meet the requirement of diversity.  

\subsubsection{Obstacle Classes and CAD Models} In order to apply for the real AD application, both the common types (such as cars, SUVs, trucks, buses, bicyclist, motorcyclist, pedestrians) and uncommon categories (e.g., traffic cone, baby carriage and tricycle) have been considered. Some examples of CAD models are shown in \figref{Fig:CAD_Modes}. The number of models for each type has been given in \tabref{tab:CAD_model_distribution}. Interestingly, a small number of 3D models can achieve high detection rate for rigid obstacles, while more models are needed for non-rigid objects. To maintain fidelity, all the 3D models are made with real size and appearance. In addition, we also try to maximize the diversity for each type as much as possible. Specifically, for vehicle models, glasses have been marked as transparent and passengers and driver are added inside them as in the real traffic scene. 
\begin{table}[h!] 
\setlength\tabcolsep{4.25pt}

\begin{tabular}{l| c c c c c c}
\scriptsize{Types} & \scriptsize{Cars, SUVs} & \scriptsize{Trucks, Buses} & \scriptsize{Bicyc- \& Motor-list}  & \scriptsize{Pedestrians} & \scriptsize{Others} \\ \hline
\scriptsize{Number} & 45 & 60 & 350 & 500 & 200
\end{tabular} 
\caption{\normalfont Number of 3D model for different categories. Some uncommon categories are included in the others such as  traffic cones, baby carriages and tricycles.}
	\label{tab:CAD_model_distribution}
\end{table} 
\vspace{-0.5cm}

\subsection {Sensor Simulation}
\label{subsec:sensor}

The LiDAR sensor captures the surrounding scene through reckoning the time of flight of laser pulses emitted from the LiDAR and reflected from target surfaces \cite{kim2013simulation}. A point is generated if the returned pulse energy is much higher than a certain threshold. 

\subsubsection{Model Design}
 \begin{figure}[!t]
 	\begin{center}
 		\includegraphics[width=0.95\linewidth]{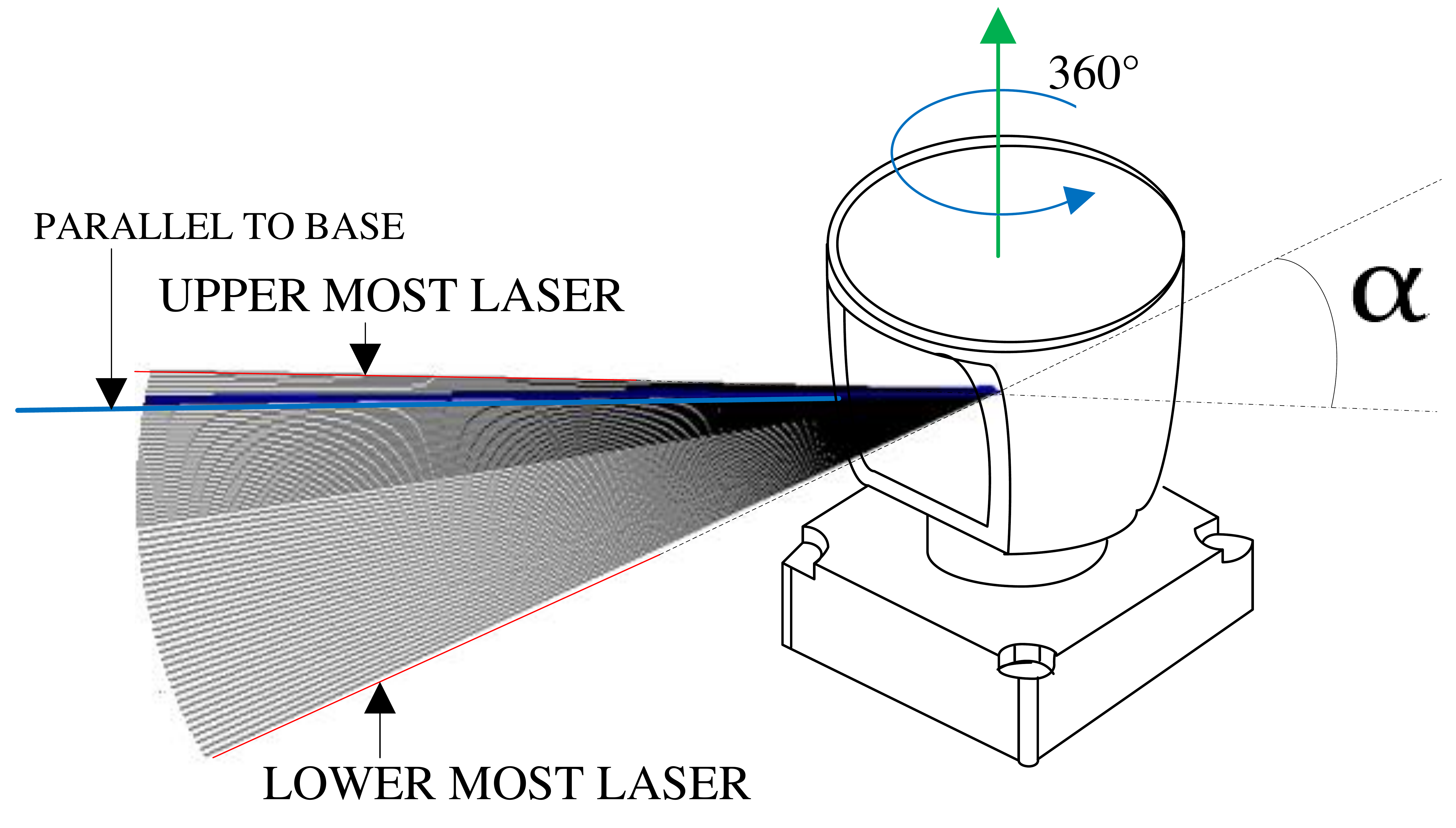}
 	\end{center}
 	\caption{Geometric model of Velodyne HDL-64E S3 which emits 64 laser beams at a preset rate and rotates to cover 360 degrees filled of view.}
 	\label{fig:vld64}
 \end{figure}
A simple but practically sufficient physical model has been used to simulate this process here. The model is formulated as
 
 
 \begin{equation}\label{eq:retrun_power}
 \renewcommand\arraystretch{1.25}
 \begin{aligned}[b]
 E_{{return}} 
 &= E_{{emit}} * R_{{rel}} * R_{{ia}} * R_{{atm}}, \\
R_{{ia}} 
&= (1-\cos\theta)^{0.5},     \\
 R_{{atm}} 
 &= exp(-\sigma_{{air}} * D),   \\
 \end{aligned}
 \end{equation}
where $E_{return}$ denotes the energy of a returned laser pulse and $E_{emit}$ is the energy of original laser pulse, $R_{rel}$ represents the reflectivity of the surface material, $R_{ia}$ denotes the reflection rate w.r.t the laser incident angle, $R_{atm}$ is the air attenuation rate because the laser beam is absorbed and reflected when traveling in the air, $\sigma_{air}$ is a constant number (\eg 0.004) in our implementation, and $D$ denotes the distance from LiDAR center to target.

Given the basic principle above, the common used multi-beam LiDAR sensor (e.g., Velodyne HDL-64E S3) for autonomous vehicles can be simulated. It emits 64 laser beams in different vertical angles ranging from \SI{-24.33}{\degree} to \SI{+2}{\degree}, as shown in \figref{fig:vld64}. These beams can be assumed to be emitted from the center of the LiDAR. During data acquisition, HDL-64E S3 rotates around its own upright direction and shoots laser beams at a predefined rate to accomplish \SI{360}{\degree} coverage of the scenes.

Theoretically, 5 parameters of the beam should be considered for generating the point cloud, including the vertical and azimuth angles and their angular noises, as well as the distance measurement noise. Ideally, these parameters should keep constant, however, we found that different devices have different vertical angles and noise. To be closer to reality, we obtain these values from real point clouds statistically.
Specifically, We collect real point clouds of these HDL-64E S3 sensors atop parked vehicles, guaranteeing the point curves generated by different laser beams to be smooth. The points of each laser beam are then marked manually and fitted by a cone with the apex located in the LiDAR center. The half-angle of the cone minus $\pi/2$ forms the real vertical angle while the noise variance is figured out from the deviation of lines constructed by the cone apex and the points from the cone surface.
The real vertical angles usually differ from the ideal ones by $1-\SI{3}{\degree}$. In our implementation, we approximate aforementioned noises using Standard Gaussian Distribution, setting distance noise variance to \SI{0.5}{\cm} and the azimuth angular noise variance \SI{0.05}{\degree}. 

\subsubsection{Point Cloud Rendering} In order to generate a point cloud, we have to compute intersections of laser beams and the virtual scene, for this we propose a cube map based method to handle the hybrid data of virtual scenes, \ie points and meshes. Instead of computing intersections of beams and the hybrid data, we compute the intersection with the  projected maps (\eg depth map) of scenes which offer the equivalent information but much easier. 

\begin{figure}[!t]
	\begin{center}
		\includegraphics[width=0.99\linewidth]{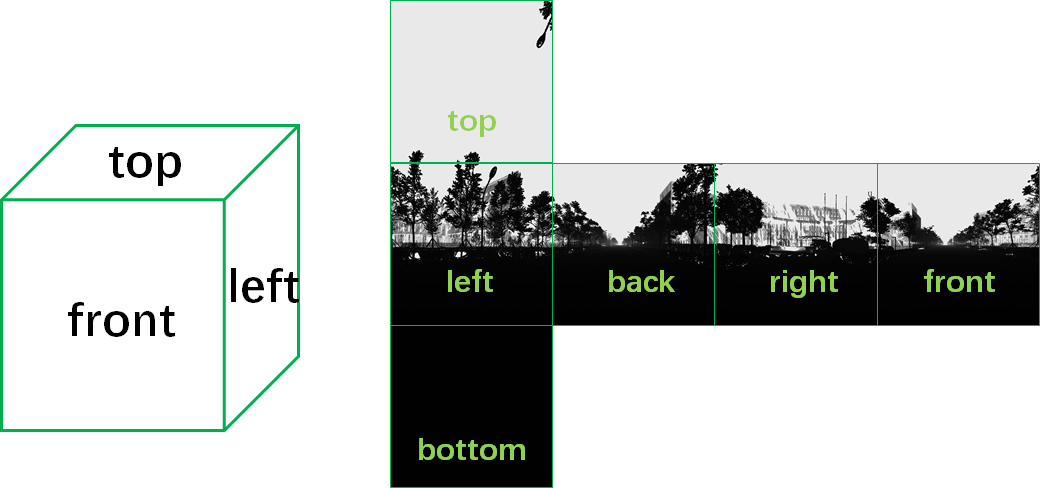}
	\end{center}
	\caption{The cube map generated by projecting the surrounding point cloud onto 6 faces of a cube centered at the LiDAR origin. Here we only show the depth maps one the 6 different views.}
	\label{fig:cubemap}
\end{figure}

To do this, we first perspectively project the scene onto 6 faces of a cube centered at the LiDAR origin to form the cube maps as in \figref{fig:cubemap}. The key to make cube maps usable is to obtain the smooth and holeless projection of the scene, with presence of environment points. Therefore we render the environment point cloud using surface splatting \cite{zwicker2001surface}, while rendering obstacle models using the regular method. We synergize these two parts in the same rendering pipeline, yielding in a complete image with both the environment and obstacles. In this way, we get 3 types of cube maps: depth, normal, and material which are used in \equref{eq:retrun_power}.

Next, we simulate the laser beams according to the geometric model of HDL-64E S3, and for each beam we look for the distance, normal and material of the target sample it hit, with which we generate a point for this beam.  Note that some beams are likely discarded, if its returned energy computed with \equref{eq:retrun_power} is too low or it hit a empty area in the cube face, which indicates the sky.

Finally, we automatically generate the tight oriented bounding box (OBB) for each obstacle by simply adjusting its original CAD' OBB to points of the obstacle.

\begin{table*}[ht!]
	\centering
		
	\begin{tabular}{rccccc}
		\Xhline{2\arrayrulewidth}
		\multirow{2}{*}{\textbf{Methods}}  & \multicolumn{2}{c}{{\textbf{Instance Segmentation}}} & \multicolumn{2}{c}{{\textbf{3D Object Detection}}} \\
		 & mean AP  & mean MasK AP  & AP 50 & AP 70   \\ \hline
		CARLA & 10.55 &  23.98 & 45.17 & 15.32     \\
		Proposed	&  \textbf{33.28} &  \textbf{44.68}   &  \textbf{66.14}   &   \textbf{29.98}             \\
		 Real KITTI	 & \B{40.22}  & \B{48.62}            &  \B{83.47} & \B{64.79}     \\ 
		\hline	
		CARLA + Real KITTI    & 40.97           & 49.31           & 84.79 & 65.85  \\  
		Proposed + Real KITTI & \textbf{45.51}  & \textbf{51.26}  & \textbf{85.42} & \textbf{71.54}       
		\\ \hline		
	\end{tabular}
	\caption{\normalfont The performance of models trained by different simulation point cloud on KITTI benchmark. In which, ``CARLA'' and ``Proposed'' represents the model trained by the point cloud generated by CARLA and proposed method, ``Real KITTI'' represents the model trained with KITTI training data and the ``CARLA + Real KITTI'' and  ``Proposed + Real KITTI'' represent the models trained on the simulation data first and the fine-tuned on the KITTI training data.}  
		\label{tab:dataset_cmp_pure_sim_semantic}
\end{table*}

\section{Experimental Results and Analysis}
\label{sec:experiments}

The whole simulation framework is a complex system. Direct comparison of different simulation system is really a difficult task and it is also not the key point of this paper. The ultimate objective of our work is to boost DNN's perception performance by introducing free auto-labeled simulation ground truth.  Therefore, the comparison of the different simulators can be transferred by comparing the point cloud generated by different ones. Here, we choose an indirect way of evaluation by comparing the DNN's performance trained with different simulation point cloud. To highlight the superiority of our proposed framework, we plan to verify it on two types of dataset: public and the self-collected point cloud. Due to the popularity of Velodyne HDL-64E in the field of AD, we set all the model parameters based on this type of LiDAR in our simulation process and all the following experiments are executed based on this type of LiDAR. 

\subsection{Evaluation on Public Dataset}

Currently, CARLA, as an open-source simulation platform \footnote{CARLA \url{ https://github.com/carla-simulator/carla}}, have been widely used for different kinds of simulation purposes (e.g.,~\cite{kaneko2018mask} \cite{yoon2018mapless} \cite{sobh2018end} \cite{kiran2018real} \cite{qiu2018deeplidar}). Similar with most typical simulation frameworks, both the foreground and background CG models have to be built in advance. As we have mentioned before, the proposed framework only need the CG models of foreground and the background can be directly obtained by a laser scanner. Here, we take CARLA as a representative of traditional methods for comparison. 

\subsubsection{Dataset}

\textsl{Simulation Data:} we use CARLA and the proposed method to produce two groups of simulation data. In order to get better generalization, we first generate 100,000 frame of point cloud from a large scale traffic scenario and then randomly select 10,000 frames for using. For CARLA, both the foreground and background point cloud are rendered simultaneously with CG models. For the proposed method, we first build a clean background and then place some certain obstacles base on the proposed PM algorithm. To be fair, similar number of obstacles are included in each group. 

\textsl{Real Data:} all the evaluation are executed on the third-party public KITTI \cite{geiger2012we}  object detection benchmark. This data has been divided into training and testing two subsets. Since the ground truth for the test set is not available, we subdivide the training data into a training set and a validation set as described in \cite{chen2016monocular} \cite{chen2017multi} \cite{zhou2018voxelnet}. Finally, we obtained 3,712 data samples for training and 3,769 data samples for validation. On the KITTI benchmark, the objects have been categorized into ``easy'', ``moderate'' and ``hard'' based on their height in the image and occlusion ratio etc. Here, we merge them together for evaluation because they are equally important for the real AD application. In addition, the intensity attribute has been removed in our experiments.  

\subsubsection{Evaluation Methods}

Two popular perception tasks in AD application have been evaluated here including instance segmentation (simultaneous 3D object detection and semantic segmentation) and 3D object detection. For each task, one state-of-the-art approach is used for evaluation here. As far, SECOND \cite{yan2018second} performs the best for 3D object detection on the KITTI benchmark among all the open-source methods. Therefore, we choose it for object detection evaluation. While for instance segmentation, an accelerated real-time version of MV3D \cite{chen2017multi} from ApolloAuto \footnote{ApolloAuto \url{ https://github.com/ApolloAuto/apollo}} project is used here. 
\subsubsection{Evaluation Metrics}

For the 3D object detection, we take the AP-50 (Average Precision) and AP-70 which have been used on the KITTI benchmark. For instance segmentation, we take the metric of mean bounding box (Bb)/mask AP proposed in coco challenges \cite{lin2014microsoft} here. Specifically, the thresholds are set as $[0.5:.05:0.95]$. 

\subsubsection{Experimental Results and Analysis}
 Here, we advocate two kinds of way for using the simulation data. One is to train a model purely using simulation data and the other is to train a model on the simulation data first and then fine-tuned on the real data. Both of them are evaluated here and the experimental results are shown in \tabref{tab:dataset_cmp_pure_sim_semantic}. Compared with CARLA, the model trained purely by the proposed simulation point cloud achieves better performances on both instance segmentation and object detection. For instance segmentation, it gives more than 20 points improvements for mean AP and Mask AP. While for object detection, the AP is also improved by a big margin. With the help of simulation data, both of the models for segmentation and object detection can be boosted compared to the model trained only on the real data. The performance has just increased slightly by CARLA data, while the mean AP for segmentation and AP 70 for object detection have been improved about 5 point by using our simulated dataset. 
 
\textsl{Analysis:} although the simulation data can really boost the DNN's performance by fine-tuning with some real data, however, its generalization ability is far from the real application. For example, the model trained only with simulation data can achieve only 29.28\% detection rate for the 3D object detection which is far from the real application in AD. The main reason of this is the domain in training is quite different with testing data and this is a very important drawback for the traditionally simulation framework. However, we would like to claim that this situation can be well recited with our proposed method. With carefully design, the model trained with simulated data can achieve applicable detection results for the real application. Detailed results will be introduced in the following subsection.     

\subsection{Simulation for Real Application}

As we have mentioned, the proposed framework can easily solve the domain problem happening in the traditional simulators by collecting the similar environment by a profession laser scanner. For building a city-level large-scale background, one week is sufficient enough. Then, we can use this background to generate sufficient ground truth data. 

To further prove the effectiveness of our proposed framework, we test it on a large self-collected dataset. We collect and label more point cloud with our self-driving car from different cities. The data is collected by Velodyne HDL-64E, including 100,000 frames for training and 20,000 for testing. Different from KIITI, our data is labeled for full \SI{360}{\degree} view. Six types of obstacles have been labeled in our dataset, including small cars, big motors (e.g., truck and bus), cyclists, pedestrians, traffic cones and other unknown obstacles. The following experiments are executed based on this dataset. 

\subsubsection{Results on Self-collected Dataset}

\begin{table}[ht!]
\centering
\begin{tabular}{R{3cm} C{2.5cm} }

	Dataset & mean Mask AP \\
	\Xhline{2\arrayrulewidth}
	100k sim & 91.02\\
    16k  real &93.27\\
    100k sim + 1.6k real & \textbf{94.10} \\
    
	\hline 
	100k real & 94.39 \\
	100k sim + 16k real & \textbf{94.41}\\
	\hline 
	100k sim + 32k real & 94.91 \\
	100k sim + 100k real & 95.27 \\
	\hline
\end{tabular}
\caption{\normalfont Model trained with pure simulation point cloud can achieve comparable results with model trained with real dataset for instance segmentation task.}
	\label{tab:boosting_real_data}
\end{table}

 First of all, we collect background point cloud from different cities to build a large background database. Then we generate sufficient simulation point cloud by placing different objects into the background. Finally, we randomly select 100K frame of simulation point cloud for our experiments. The experimental results are shown in \tabref{tab:boosting_real_data}. From the first big row of the table, we can find that the model trained with 100K simulation data gives comparable result with the model trained with 16K real data, with only 2 points of gap. More important, the detection rate can achieve 91.02\% which is relatively high enough for the real application. Furthermore, by mixing 1.6K real data with the 100K simulation data, the mane AP reaches $94.10$ which outperforms 16K real data.
 
 From the second row of the table, we can see that 16K real data together with 100K simulation can beat 100K real data, which can save more than 80\% of the money for annotation. Furthermore, we can obviously found last row the table that model trained with real data can be boosted to varying degrees by adding simulation data even we have big enough labeled real data. 

\subsubsection{Ablation Studies}
The results in \tabref{tab:boosting_real_data}, shows the promising performance of the proposed simulation framework. While the whole simulation framework is a complex system and the final perception results comes from the influence of different steps of the system. Inspired by the idea of ablation analysis, a set of experiments have been designed to learn the effectiveness of different parts. Generally, we found that three main parts are very important for the whole system including the way of background construction, obstacle pose generation, random point dropout. Detailed information for each part will be introduced in this section.

\textsl{Background}
As discovered by many researchers, context information from the background is very important for perception task. As shown in \tabref{tab:dataset_cmp_background}, we have found that the background simulated from our scanned point cloud (bold values) can achieve very close performance with the real background (values in blue). Compared with synthetic models (e.g., CARLA), the performance can be improved about $10$ points for both mean Bbox and mask APs.
\begin{table}[t]
	\centering
	\small
	\begin{tabular}{rcccccccc}
		\Xhline{2\arrayrulewidth}
		Methods &  Mask AP 50 & Mask AP 70 & mean Mask AP \\
		\hline
		 No BG  + Sim FG    &  1.86 & 1.41 & 1.25 \\
		Scan BG + Sim FG   &  \textbf{88.60 } & \textbf{86.80 } &  \textbf{83.38}\\
		\hline
		Real BG + Sim FG   &  88.79 &  87.19 & 84.22\\
		Real BG + Real FG  & \B{90.40} &  \B{89.45}  &\B{86.33}\\
		\hline
	\end{tabular}
	\caption{\normalfont Evaluations with different background for instance segmentation, where ``Sim'', ``BG'' and ``FG'' represent ``simulation'',``background'' and ``foreground'' for short.}
	
	\label{tab:dataset_cmp_background}
\end{table}

 \begin{table}[ht!]
	\small
	\centering
	
	\begin{tabular}{rcccccccc}
		\Xhline{2\arrayrulewidth}
		Methods &  Mask AP 50 & Mask AP 70 & mean Mask AP \\
		\hline
		Random on road &73.03   &69.23  & 65.95\\
		Rule-based   & 81.37  & 78.13  &74.32\\
		Proposed PM  & 86.57 & 84.55 & 82.47\\
		Augmented PM & \textbf{87.80 }& \textbf{86.19} & \textbf{83.07}\\
		\hline
		Real Pose   &  \B{88.60 } & \B{86.80 }&\B{83.38}\\
		
		\hline
	\end{tabular}
	\caption{\normalfont Evaluations for instance segmentation with different obstacle poses.}
	\label{tab:dataset_cmp_pose}
\end{table}
\textsl{Obstacle Poses} We also empirically find that where to place objects into background has a big influence on the perception results. Five different ways of placing obstacles have been evaluated here: (a) randomly on the road, (b) rule based method depends on prior high definition map information, (c) proposed probability map (PM) based method, (d) probability map plus some pose augmentation, (e) manual labeled pose plus data augmentation. From \tabref{tab:dataset_cmp_pose}, we can find that the proposed PM method together with proper pose augmentation can achieve comparable results with the manual labeled real pose while the proposed method can largely free the manual labors.

\textsl{Random Dropout}
Another difference between the real and simulated data is the number of the point in each frame. For Velodyne HDL-64E,  usually about 102,000 points will be returned each frame for the real sensor, while the number is about 117,000 in the simulated point cloud. The reason for this may come from different aspects: textures, material, surface reflectivity and even the color of objects. For man-made foreground obstacle, these proprieties can be easily obtained, however, it is not a trail task for backgrounds. Inspired by the dropout strategy in DNN, we randomly drop a certain ratio of points during our model training process. Surprising, we find that it can steadily improve the performance by 2 points for  mean Mask AP.       
 \begin{table}[ht!]
	\small
	\centering
	\begin{tabular}{rcccccccc}
		\Xhline{2\arrayrulewidth}
		Methods &  Mask AP 50 & Mask AP 70  & mean Mask AP \\
		\hline
		w/o dropout   & 88.60 & 86.80 & 83.38 \\  
		w dropout      &  \textbf{89.43} &  \textbf{88.75} &  \textbf{85.53} \\ 
		\hline
		Real Data    &  \B{90.40} &  \B{89.45} & \B{86.33}\\ 	
		\hline
	\end{tabular}
	\caption{\normalfont Evaluations for instance segmentation with or without random dropout.}
	\label{tab:random_dropout}
\end{table}

\subsubsection{Extension to Different Sensors}
\begin{figure}[htbp]
	\begin{center}
		\includegraphics[width=0.50\textwidth]{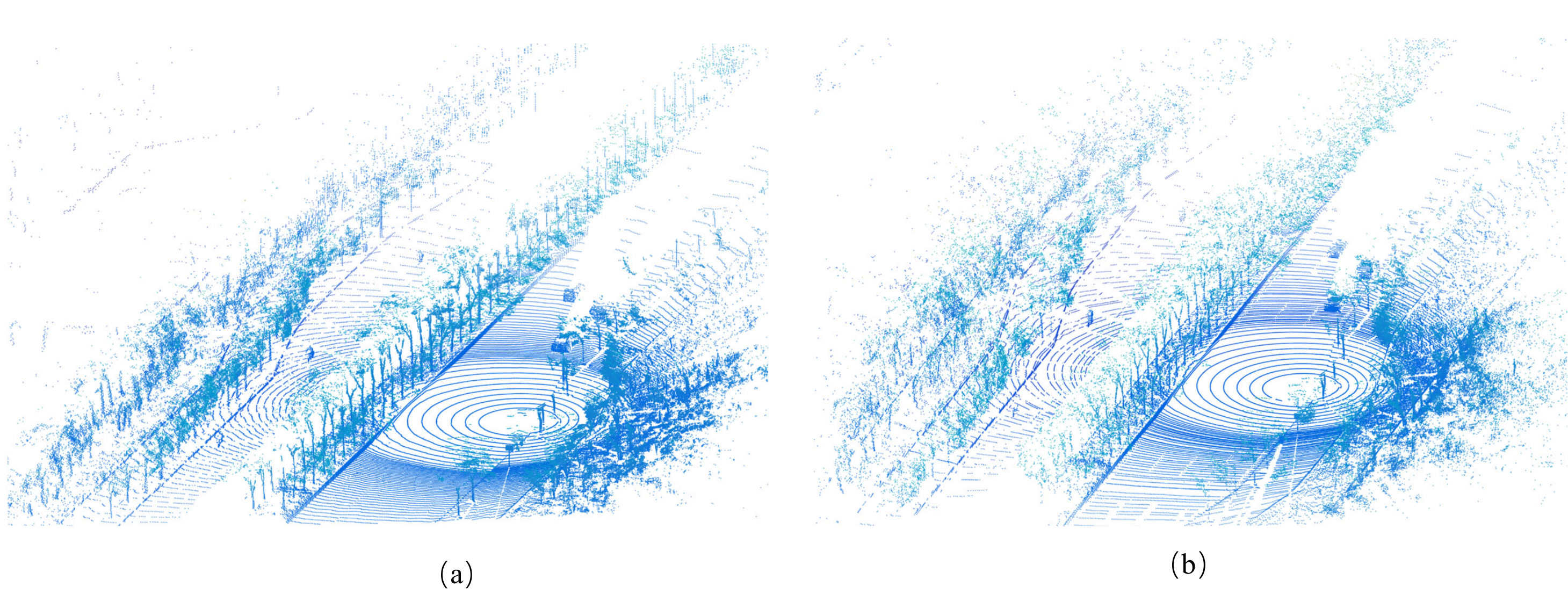}
	\end{center}
	\caption{An simulation example of VLS-128 with same LiDAR location and obstacle posse, where (a) is the simulated point cloud and (b) is the real one.}
	\label{fig:128-sim}
\end{figure}

Beyond the Velodyne HDL-64E LiDAR, the proposed framework can be easily generalized to other type sensors such as  Velodyne Alpha Puck (VLS-128) \footnote{Velodyne Alpha (VLS-128) \url{https://velodynelidar.com/vls-128.html}}. For our implementation, a friendly interface have been designed. With a slightly modification of configure file, we can achieve different types of LiDAR point clouds. In the configure file, the specific LiDAR properties can be modified such as the channels number, range, horizontal and vertical filed-of-view, angular and vertical resolution etc. \figref{fig:128-sim} gives a simulation example of VLS-128. By using the same LiDAR location and obstacle poses, we can't visually distinguish the real and simulated one.  


\section{Conclusions and Future Works}
\label{sec:conclusions}
This paper presents an augmented LiDAR simulation system to automatically produce annotated point cloud used for 3D obstacle perception in autonomous driving scenario. Our approach is entirely data driven, with scanned background, obstacle poses and types that are statistically similar to that from real traffic data, and a general LiDAR renderer that takes into considerations of physical/statics properties of the actual devices. 
The most significant benefits of our approach are realism and scalablity. We demonstrated realism by showing that the performance gap between detectors trained with real or simulated data is within two percentage point. The scalablity of our system is by design, there is no manual labeling, and the combination of different background and different obstacle placement provides abundant labeled data with virtually no cost except computation. More importantly, it can rapidly simulate a scene of interests by simply scanning that area.


Looking into the future, we want to investigate the use of low-fidelity LiDAR. Our current hardware system can produce very high quality and dense 3D point cloud, which can be re-sampled to simulate any LiDAR type. 
By using a low-fidelity one, we think it may limit the LiDAR type our system can simulate, this could be a design choice between cost and versatility. Another area to address is to simulate the LiDAR intensity value for the foreground object. It involves the modeling of material reflectance properties under near infrared (NIR) illumination, which is doable but quite tedious. CG models for foreground needs to be updated accordingly to include an NIR texture. 

{
\small
\bibliographystyle{ieee}
\bibliography{references}
}

\end{document}